\ifdefined\pdfminorversion
\fi
\documentclass[letterpaper,journal]{IEEEtran}

\usepackage{amsmath,amsfonts,amssymb}
\usepackage{array}
\usepackage[caption=false,font=footnotesize,labelfont=sf,textfont=sf]{subfig}
\usepackage{textcomp}
\usepackage{stfloats}
\usepackage{url}
\usepackage{graphicx}
\usepackage{booktabs}
\usepackage{multirow}
\usepackage{xcolor}
\usepackage{cite}
\usepackage{algorithm}
\usepackage{algpseudocode}
\usepackage{balance}
\usepackage[colorlinks=true, linkcolor=blue, citecolor=blue, urlcolor=black]{hyperref}

\graphicspath{{figures/figure_v2/}}

\begin{document}

\title{RAPAC-DP: Response-Aligned Pending-Action Compensation for Diffusion Policies under Delayed Execution}

\author{Tao Wang, Wei Wang, Jianhui Wang, Zhengyang Ye, Qi Wang, Weidi Huang, and Bing Xu%
\thanks{This work was supported by the State Key Laboratory of High-end Heavy-load Robots (Open Fund Project No. HHR2025020105).}%
\thanks{Tao Wang, Jianhui Wang, Qi Wang, Zhengyang Ye, Weidi Huang, and Bing Xu are with Zhejiang University, Hangzhou, Zhejiang, China.}%
\thanks{Wei Wang is with the State Key Laboratory of High-end Heavy-load Robots, Midea Group, Foshan, Guangdong, China.}%
\thanks{Corresponding authors: Weidi Huang (e-mail: wdhuang@zju.edu.cn) and Wei Wang (e-mail: wangwei232@midea.com).}%
}

\markboth{Preprint}
{Wang \MakeLowercase{\textit{et al.}}: RAPAC-DP for Diffusion Policies under Delayed Execution}

\maketitle

\begin{abstract}
Cloud-side inference expands the computational resources available to imitation-learning policies, but communication and inference latencies create a mismatch between prediction and execution. We propose RAPAC-DP, a response-aligned pending-action compensation framework for diffusion- and flow-based action generators. By conditioning on actions committed during inference, RAPAC-DP generates a full-horizon action chunk aligned with the scheduled handover time. A lightweight pending-action encoder and LoRA adapters learn this compensation while keeping the pretrained base policy frozen. We construct delay-conditioned training samples from delay-free demonstrations, requiring neither an explicit dynamics model nor additional data collected under delayed execution. In simulation benchmarks at the largest evaluated delays, RAPAC-DP retains 81.4\% of its overall delay-free performance on Kinetix and achieves a mean success rate of 63.3\% across three RoboMimic tasks. Across 450 physical trials on a Franka Panda manipulator, RAPAC-DP increases the high-delay mean success rate from 28.9\% for RTC to 60.0\% across static and conveyor-based pick-and-place tasks.
\end{abstract}

\section{Introduction}

\IEEEPARstart{L}{earning} from demonstration has supported smooth trajectory learning and task-oriented motion assistance in robotic manipulation \cite{LfD,luo2025trajectory,cheng2024virtual}. Deep imitation-learning policies extend this paradigm to complex visuomotor tasks by mapping high-dimensional observations to actions \cite{diffusion_policy,act,bet,rt1}. As policy models grow in size and complexity, however, inference can require more computing power than many robotic platforms can provide locally \cite{rt2,palme}. To meet these demands, policy inference can be performed in the cloud, where greater computational resources are available. In this architecture, the robot transmits observations to a cloud server and executes the actions returned by the policy \cite{fogros2,network_offloading}. Cloud inference reduces the computational burden on the robot, but communication and remote computation introduce delays into the control loop.

\begin{figure}[!t]
\centering
\includegraphics[width=\columnwidth]{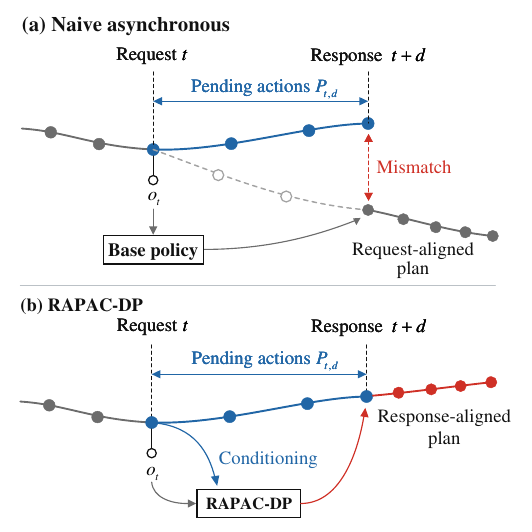}
\caption{Schematic of observation--execution alignment under execution delay. (a) Naive inserts the complete request-aligned chunk at $t+d$, beginning with $\hat{\mathbf{a}}_t^{(t)}$. The pending actions $\mathbf{P}_{t,d}$ change the robot state during inference, producing an observation--execution mismatch. (b) RAPAC-DP conditions on $\mathbf{P}_{t,d}$ and generates a complete action chunk beginning at the scheduled handover $t+d$.}
\label{fig:latency-misalignment}
\end{figure}

This end-to-end control latency stems from observation upload, cloud-side model inference, and control-command transmission \cite{network_offloading}. Its duration varies with computational workload and network conditions, while execution jitter and asynchronous control updates introduce additional timing uncertainty \cite{async_rl_physical_robots,fogros2_plr}. As illustrated in Fig.~\ref{fig:latency-misalignment}, when a delayed action chunk arrives, a naive asynchronous executor inserts the complete returned chunk at the queue head and begins executing its first action. The chunk was generated from the request-time observation, without accounting for the pending actions executed during inference. This mismatch breaks the observation--action alignment assumed during delay-free policy training and makes the observed control process non-Markovian \cite{timedelayed_mdp}. Consequently, policies trained under delay-free conditions may suffer performance degradation or instability when deployed in cloud-based control systems.

Delayed control has traditionally been addressed through information-state representations and model-based state prediction. Information-state methods augment the current observation with action or observation histories, but longer delays require longer histories, increasing both input dimensionality and the difficulty of policy learning \cite{state_augmentation,transformer_delay_rl}. Model-based methods instead predict the state reached during the delay interval, making their performance dependent on model accuracy and multi-step prediction quality \cite{delayaware_mbrl,delayed_obs_world_models}. More recent asynchronous action-chunking methods overlap inference with execution and coordinate successive chunks through inpainting, online refinement, or masked generation \cite{rtc,a2c2,remac}. These approaches highlight the importance of accounting for both changes in the robot state and the actions executed while the new plan is being generated.

During cloud inference, the local controller continues executing queued actions, which we refer to as the pending-action sequence. These actions change the robot's state before the new prediction arrives. A scalar delay specifies the elapsed number of control steps but does not identify the actions executed during that interval. RAPAC-DP therefore conditions on the pending-action sequence and generates an action chunk aligned with the response arrival time.

We propose RAPAC-DP, a response-aligned pending-action compensation framework for cloud-deployed action-generation policies. RAPAC-DP constructs training samples directly from delay-free expert trajectories. For each sample, the actions executed during a selected delay interval serve as the pending-action input, while the subsequent actions provide the response-aligned target. This construction requires no additional demonstrations under delayed execution. For parameter-efficient adaptation, RAPAC-DP adds a pending-action encoder and LoRA adapters to a frozen base policy and trains only these added modules. These added modules can be bypassed to recover the frozen base policy.

The main contributions of this paper are summarized as follows:
\begin{itemize}
\item We introduce a training-sample construction strategy that converts delay-free expert trajectories into samples for delayed execution. The actions executed while the robot waits for the cloud response are used as pending-action inputs, and each target contains a complete action sequence beginning when the response arrives. This construction enables response-aligned training without requiring additional demonstrations under delayed execution.

\item We develop RAPAC-DP by adding a pending-action encoder and LoRA adapters to a frozen action-generation policy. Only the added modules are trained for delay compensation. The compensation pathway can be bypassed to recover the frozen base policy.

\item We evaluate RAPAC-DP across three RoboMimic tasks, 12 Kinetix environments, and three physical pick-and-place tasks on a Franka Panda manipulator. At the largest evaluated delays, RAPAC-DP retains 81.4\% of its delay-free performance on Kinetix and achieves a mean success rate of 63.3\% on RoboMimic. Across 450 physical trials, it increases the high-delay mean success rate from 28.9\% for RTC to 60.0\%.
\end{itemize}

\section{Related Work}
\label{sec:related_work}

\subsection{Diffusion- and Flow-Based Action Generation}
Diffusion models learn complex distributions through iterative denoising and have demonstrated strong generative performance in image synthesis \cite{ddpm,improved_ddpm}. They have since been extended to planning and offline reinforcement learning, where they model trajectory and action distributions \cite{diffuser,diffusion_offline_rl}. In robot manipulation, Diffusion Policy formulates visuomotor control as conditional denoising over multi-step action sequences and executes only the near-term portion of each sequence before replanning \cite{diffusion_policy}. Subsequent work reduces generation cost through consistency distillation and consistency flow matching \cite{consistency_policy,flowpolicy}. These methods improve action generation and sampling efficiency but do not address the prediction--execution offset introduced by delayed inference.

\subsection{Latency Compensation and Asynchronous Execution}
Delayed decision-making has been studied through time-delayed MDPs and augmented information states that combine delayed observations with intervening actions \cite{timedelayed_mdp,state_augmentation}. Transformer- and encoder-based methods instead learn from histories of delayed states and intervening actions \cite{transformer_delay_rl,deer}. Other approaches incorporate delay into learned dynamics, infer current information with world models, resample delayed trajectory fragments, or introduce temporal skip connections \cite{delayaware_mbrl,delayed_obs_world_models,random_delays,handling_delay_rl}. Adversarial imitation learning has also been used to learn delay-robust policies from undelayed demonstrations \cite{delayed_adversarial_imitation}.

Recent studies focus on the asynchronous execution of action-chunking policies. Real-Time Chunking, RTC~\cite{rtc}, preserves committed actions and guides the generation of the remaining chunk during inference. A2C2 applies per-step corrective updates to base actions using the latest observation~\cite{a2c2}. REMAC masks the committed action prefix during training and holds it fixed during sampling to maintain inter-chunk continuity~\cite{remac}. At the architecture level, SmolVLA decouples prediction from execution, whereas VLA-RAIL smooths and blends successive chunks~\cite{smolvla,vlarail}. VLASH rolls the robot state forward under scheduled actions and trains the policy with temporally offset targets~\cite{vlash}.

\subsection{Parameter-Efficient Policy Adaptation}
Low-Rank Adaptation (LoRA) represents updates to frozen weights with trainable low-rank factors, reducing the number of trainable parameters and adapter storage overhead \cite{lora}. In robot learning, LoRA has enabled efficient adaptation of Vision-Language-Action (VLA) policies to new tasks and robot embodiments \cite{openvla,lorasp}. REMAC also uses LoRA to adapt a pretrained action-generation policy for asynchronous execution \cite{remac}.
\section{Preliminaries}

\subsection{Generative Action-Sequence Policies}

Let $\mathbf{x}^0$ denote a normalized action chunk. Diffusion-based policies perturb $\mathbf{x}^0$ across discrete generative steps $s$ according to
\begin{equation}
\mathbf{x}^s = \sqrt{\bar{\alpha}^s}\,\mathbf{x}^0 + \sqrt{1-\bar{\alpha}^s}\,\boldsymbol{\epsilon},
\qquad \boldsymbol{\epsilon}\sim\mathcal{N}(\mathbf{0},\mathbf{I}),
\label{eq:diffusion-closed-form}
\end{equation}
where $\bar{\alpha}^s$ denotes the cumulative noise schedule~\cite{ddpm}. A conditional reverse denoising process maps noisy samples back to clean action chunks conditioned on observation $\mathbf{o}$~\cite{diffusion_policy}.

Flow-based policies generate the action chunk by integrating an observation-conditioned velocity field~\cite{flowpolicy}:
\begin{equation}
\frac{d\mathbf{x}^{\tau}}{d\tau} = \mathbf{v}_{\theta}(\mathbf{x}^{\tau},\tau \mid \mathbf{o}).
\label{eq:flow-ode}
\end{equation}
The generative indices $s$ and $\tau$ represent sampling time and remain distinct from the physical control step $t$.


\subsection{Low-Rank Adaptation}
\label{sec:prelim_lora}

Low-Rank Adaptation (LoRA) is a parameter-efficient method for adapting pretrained models by introducing low-rank updates to frozen weights. Given a pretrained weight matrix $W_0 \in \mathbb{R}^{m \times n}$, LoRA models the update as $\Delta W = \frac{\alpha_L}{r}BA$, where $A \in \mathbb{R}^{r \times n}$ and $B \in \mathbb{R}^{m \times r}$ are trainable low-rank matrices with $r \ll \min(m,n)$. The forward pass is given by
\begin{equation}
    h = W_0 x + \Delta Wx.
\end{equation}
In this formulation, $W_0$ is frozen and only the low-rank parameters are optimized.

\subsection{Control under Execution Delay}

Execution delay $d$ is the number of control steps from sending observation $\mathbf{o}_t$ to executing the returned action chunk. Early responses are buffered until this scheduled step.

\paragraph{No-delay case}
In the delay-free setting, the policy receives observation $\mathbf{o}_t$ at time $t$ and predicts an $H$-step action sequence, where $H$ denotes the prediction horizon:
\begin{equation}
\hat{\mathbf{A}}^{(t)}
=
\pi(\mathbf{o}_t)
=
\left[
\hat{\mathbf{a}}_t^{(t)},
\hat{\mathbf{a}}_{t+1}^{(t)},
\dots,
\hat{\mathbf{a}}_{t+H-1}^{(t)}
\right].
\label{eq:plan-no-delay}
\end{equation}
The controller executes the corresponding actions immediately. The parameter $K$ denotes the query interval in control steps, bounded by $1 \leq K \leq H$, such that the controller executes the first $K$ actions before issuing the next policy request at time $t+K$:
\begin{equation}
\mathbf{a}_k^{\text{exec}} = \hat{\mathbf{a}}_k^{(t)}, \quad \forall k \in [t, t+K-1].
\label{eq:execute-no-delay}
\end{equation}

\paragraph{Naive asynchronous execution under delay}
During remote inference, the robot continues executing the committed pending actions $\mathbf{P}_{t,d}$ from its local queue. At control step $t+d$, the Naive executor inserts the complete prediction $\hat{\mathbf{A}}^{(t)}=\pi(\mathbf{o}_t)$ at the queue head, so that its first action $\hat{\mathbf{a}}_t^{(t)}$ is executed at step $t+d$. This chunk is generated from the request-time observation without conditioning on the intervening pending actions. In contrast, RAPAC-DP conditions on those actions and predicts a complete $H$-step chunk beginning at the scheduled handover.


\begin{figure*}[!t]
\centering
\includegraphics[width=\textwidth]{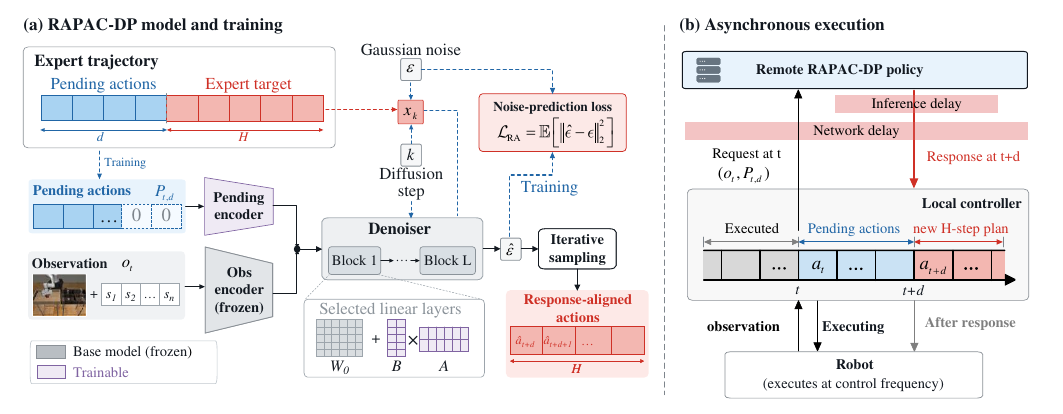}
\caption{Overview of the RAPAC-DP framework. (a) Response-aligned training on a diffusion policy. Delay-free demonstrations provide pending-action conditions and full-horizon targets beginning at the sampled delay step. Only the pending-action encoder and LoRA adapters are updated while the base policy remains frozen. (b) Asynchronous online execution. During remote inference, the local controller executes committed pending actions and inserts the newly received action chunk into the execution queue at the scheduled handover time.}
\label{fig:overview}
\end{figure*}

\section{Method}

\subsection{Delay-Conditioned Training Sample Construction}

We construct training samples directly from delay-free expert trajectories. Let $\xi=\{(\mathbf{o}_i,\mathbf{a}_i)\}_{i=1}^{T}$ denote an expert trajectory. For a sampled delay $d$, the start step $t$ is selected to ensure that both the pending actions and the full $H$-step target chunk fall within the trajectory length. The corresponding pending-action sequence and response-aligned target are then extracted from the same trajectory. Here, $d_{\max}$ denotes the maximum training delay. The delay ranges and sampling schedules used for each benchmark are detailed in Section V-B.

Rather than training the policy on request-aligned targets $\mathbf{a}_t$ through $\mathbf{a}_{t+H-1}$, we shift the supervision window by $d$ steps to align it with the response time $t+d$. For a sampled delay $d$, we construct
\begin{equation}
\begin{aligned}
\mathbf{A}^{\mathrm{tar}}_{t,d}
&= [\mathbf{a}_{t+d},\mathbf{a}_{t+d+1},\ldots,\mathbf{a}_{t+d+H-1}],\\
\mathbf{A}^{\mathrm{pend}}_{t,d}
&= [\mathbf{a}_{t},\mathbf{a}_{t+1},\ldots,\mathbf{a}_{t+d-1}].
\end{aligned}
\end{equation}
The target $\mathbf{A}^{\mathrm{tar}}_{t,d}$ always contains $H$ actions, preserving the full prediction horizon. The pending-action sequence contains the actions that would be executed locally while cloud inference is in progress. The sequence is empty when no delay is present. These pending actions serve as the conditioning signal (Section~\ref{sec:pending_action_encoder}) and are excluded from the target.

\subsection{Pending-Action Encoder}
\label{sec:pending_action_encoder}

The pending-action sequence specifies the actions committed before handover. For batch training, each sequence $\mathbf{A}^{\mathrm{pend}}_{t,d}$ is padded to $d_{\max}$ slots, with a binary validity mask identifying valid actions:
\begin{equation}
m^{(d)}_j = \mathbb{I}[j<d], \qquad j=0,\ldots,d_{\max}-1.
\end{equation}

Valid pending actions are normalized with the base policy normalizer, while padded slots are masked out. The normalized sequence $\bar{\mathbf{A}}^{\mathrm{pend}}_{t,d}$ is encoded via
\begin{equation}
\mathbf{z}^{\mathrm{pend}}_{t,d}
=
g_\phi\left(
\bar{\mathbf{A}}^{\mathrm{pend}}_{t,d},
\mathbf{m}^{(d)}
\right),
\end{equation}
where encoder $g_\phi$ applies mask $\mathbf{m}^{(d)}$ to exclude padding from the conditioning representation.

\subsection{Response-Aligned Policy Training}

Using the delay-conditioned samples constructed above, we train the policy to predict the response-aligned action target conditioned on the request-time observation and pending-action representation. We retain the original training objective of the base policy and apply it to the shifted target:

\begin{equation}
\mathcal{L}_{\mathrm{RA}}
=
\mathbb{E}_{t,d}
\left[
\ell_{\mathrm{base}}
\left(
\mathbf{A}^{\mathrm{tar}}_{t,d}
\mid
\mathbf{o}_t,
\mathbf{z}^{\mathrm{pend}}_{t,d}
\right)
\right],
\end{equation}
where $\ell_{\mathrm{base}}$ denotes the original training objective of the base action-generation policy. RAPAC-DP changes the conditioning input and target timing without altering the generative loss, allowing the same formulation to be used with both standard diffusion and flow-based action-generation policies.

For parameter-efficient training, the pretrained base policy is frozen while LoRA adapters are integrated into selected linear projections of the policy backbone. Together with the pending-action encoder, these modules form the trainable compensation pathway, restricting optimization strictly to the newly added parameters. The compensation pathway can be explicitly bypassed to recover the original base policy exactly under matched observations and sampling noise.

\subsection{Asynchronous Policy Execution}

During deployment, the controller extracts pending actions directly from the local execution queue. For a policy query issued at control step $t$, let $d$ denote the scheduled handover interval, which is constant in simulation and calibrated in physical deployment. The controller commits to the corresponding $d$ queued actions:
\begin{equation}
\mathbf{P}_{t,d}
=
[
\mathbf{a}^{\mathrm{queue}}_{t},
\mathbf{a}^{\mathrm{queue}}_{t+1},
\ldots,
\mathbf{a}^{\mathrm{queue}}_{t+d-1}
].
\end{equation}
These queued actions are normalized and masked as in training, then executed by the local controller while remote inference proceeds.

Conditioned on the request-time observation and pending-action sequence, the policy predicts a response-aligned action chunk
\begin{equation}
\hat{\mathbf{A}}_{t,d}
=
[
\hat{\mathbf{a}}_{t+d},
\hat{\mathbf{a}}_{t+d+1},
\ldots,
\hat{\mathbf{a}}_{t+d+H-1}
].
\end{equation}
By the time the response arrives at $t+d$, the $d$ pending actions have already been executed and consumed by the local controller. The returned chunk is therefore inserted at the current queue head, providing $H$ fresh actions starting from $t+d$. Because the prediction is aligned with the response time, the full predicted action chunk remains available for subsequent execution.

\section{Experiments}

We evaluate RAPAC-DP under increasing fixed execution delays on RoboMimic's Can, Square, and Transport manipulation tasks and on 12 Kinetix 2D physics-based control tasks. We additionally evaluate physical pick-and-place performance on a stationary cube, a plastic eggplant, and a cube moving on a conveyor belt.

\subsection{Simulation Environments and Datasets}

\paragraph{RoboMimic}
We evaluate the proposed method on the proficient-human datasets of three RoboMimic tasks~\cite{mandlekar2021matters}: \textit{Can}, \textit{Transport}, and \textit{Square}. These tasks cover object transfer, coordinated transport, and precision insertion, respectively. The policy receives visual observations and robot proprioceptive states, and performance is measured by task success rate.

\paragraph{Kinetix}
We further evaluate the proposed method on 12 human-designed environments from Kinetix~\cite{matthews2024kinetix}, covering diverse 2D physics-based control tasks such as locomotion, balancing, manipulation, and object interaction. Expert demonstrations are used for policy training, and performance is evaluated by task success rate.

\subsection{Evaluation Methodology}
For the simulation benchmarks, we emulate execution latency by injecting a fixed delay of $d$ control steps into the policy--environment loop.

\subsubsection{Comparison Paradigms}

For the simulation benchmarks, we compare four compensation paradigms under delayed execution:
\begin{itemize}
\item \textbf{Naive asynchronous baseline:} the frozen base policy is executed asynchronously under the injected delay, inserting its complete predicted chunk at the queue head at step $t+d$, starting with $\hat{\mathbf{a}}_t^{(t)}$. This baseline uses neither inference-time guidance nor trainable latency compensation.
\item \textbf{RTC:} Real-Time Chunking applies inference-time guidance to coordinate previously committed actions with the newly generated action chunk~\cite{rtc}. It operates on the frozen base policy and introduces no additional trainable parameters.
\item \textbf{A2C2:} benchmark-specific residual correction policies correct nominal base actions using recent observations~\cite{a2c2}.
\item \textbf{RAPAC-DP, proposed:} a pending-action encoder and LoRA adapters adapt the frozen base policy for response-aligned generation. The compensation pathway can be explicitly bypassed to recover the original base policy.
\end{itemize}

\subsubsection{Training and Evaluation Protocol}

Simulation benchmarks evaluate 50 rollouts per condition across random seeds under fixed execution delays, reporting the mean and standard deviation. RoboMimic evaluates delays from $d=0$ to $d=7$ with a 16-step prediction horizon, while Kinetix evaluates delays from $d=0$ to $d=4$ with an 8-step horizon. At zero delay where $d=0$, RAPAC-DP keeps the compensation pathway enabled with an empty pending-action condition, evaluating the complete model at the boundary of the delay range, while Naive provides the delay-free base policy baseline.

\paragraph{Implementation Details}
For Diffusion Transformer policies, each normalized pending action and its validity indicator are mapped to a 256-dimensional token through a two-layer MLP with Mish activations. Learned positional embeddings are added before appending these tokens to the condition-memory sequence, where invalid entries are masked out during attention. LoRA adapters are integrated into the linear projections of attention and feedforward blocks, trained with the AdamW optimizer using a constant learning rate.

For Kinetix FlowPolicy, each pending action is projected into a 64-dimensional slot feature. The temporally ordered features are then flattened and mapped to an $H\times256$ conditioning feature, which is added residually to the policy input projection.

All three RoboMimic tasks use LoRA rank $r=64$ and scaling factor $\alpha_L=64$, whereas Kinetix uses $r=16$ and $\alpha_L=16$, both with a dropout rate of 0.05. Across both benchmarks, the trainable compensation pathway introduces minimal overhead, requiring only 2.6\% and 9.0\% of the frozen base model parameters on RoboMimic and Kinetix, respectively.

\subsection{Performance under Increasing Execution Delay}

\begin{figure*}[!t]
\centering
\includegraphics[width=\textwidth]{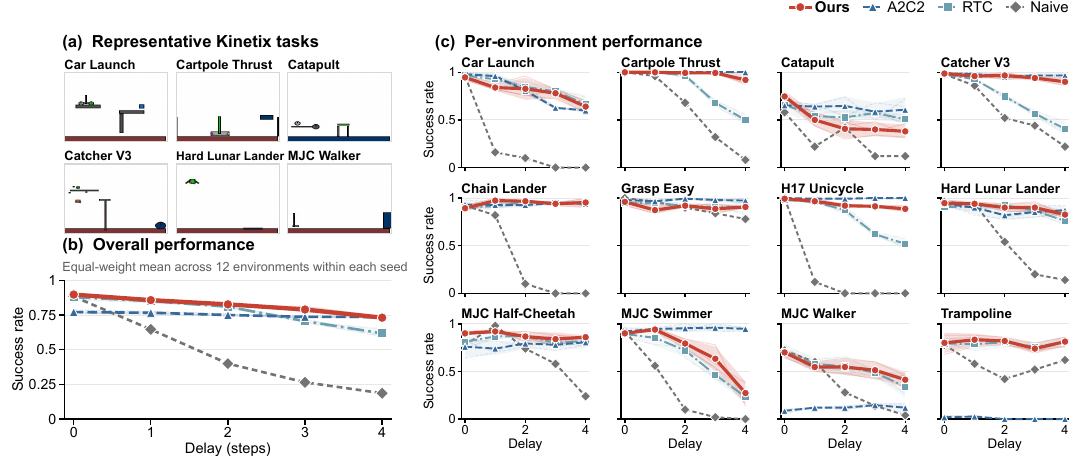}
\caption{Success rates under increasing execution delay on 12 Kinetix tasks. (a) Representative environments. (b) Overall performance, obtained by averaging the 12 environments equally within each seed before computing the across-seed mean. (c) Per-environment performance. RAPAC-DP keeps the compensation pathway enabled at every delay level, including $d=0$. For RAPAC-DP, A2C2, and RTC, thick curves show the mean, shaded bands show the standard deviation across three seeds, and pale curves show individual seeds. Naive is evaluated with one seed and therefore has no uncertainty band.}
\label{fig:delay_robustness_kinetix}
\end{figure*}

\begin{figure*}[!t]
\centering
\includegraphics[width=\textwidth]{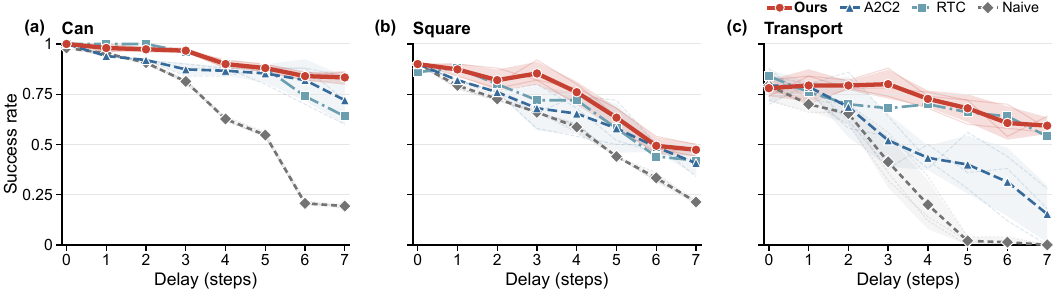}
\caption{Success rates under increasing execution delay on RoboMimic: (a) Can, (b) Square, and (c) Transport. Ours denotes RAPAC-DP. For RAPAC-DP, A2C2, and Naive, thick curves and shaded bands denote the mean and sample standard deviation across three random seeds, while pale curves show individual runs. RTC reflects the success rate from 50 rollouts per condition.}
\label{fig:delay_robustness_robomimic}
\end{figure*}

Figures~\ref{fig:delay_robustness_kinetix} and~\ref{fig:delay_robustness_robomimic} report success rates under increasing execution delays. Under the maximum evaluated delay levels, RAPAC-DP achieves a mean success rate of 63.3\% on RoboMimic at $d=7$, while retaining 81.4\% of its delay-free performance on Kinetix at $d=4$.

On Kinetix, RAPAC-DP maintained strong performance across the evaluated delay range. With the compensation pathway enabled and the pending-action condition empty, RAPAC-DP achieved an equal-weight mean of 0.898 across the 12 environments at $d=0$, slightly above the 0.883 of the Naive base policy. At $d=4$, its mean remained 0.731, retaining 81.4\% of its delay-free performance, whereas Naive retained only 21.2\%. At this delay, RAPAC-DP outperformed RTC and Naive by 11.3 and 54.4 percentage points, respectively. This aggregate robustness was reflected across distinct control dynamics, including scores of 0.953 on Chain Lander, 0.920 on Cartpole Thrust, and 0.887 on H17 Unicycle.

On RoboMimic under the maximum delay of $d=7$, RAPAC-DP achieves success rates of 83.3\% on Can, 47.3\% on Square, and 59.3\% on Transport, yielding a cross-task mean of 63.3\%. Under identical conditions, RTC achieves 64.0\%, 42.0\%, and 54.0\%, respectively, with a mean of 53.3\%, while Naive fails entirely on Transport. RAPAC-DP thus secures the highest performance across all three tasks, maintaining a 10.0-percentage-point lead over RTC in the cross-task average.

Together, the Kinetix and RoboMimic results show that RAPAC-DP maintained strong performance at $d=0$ and remained robust as execution delay increased.

\subsection{Ablation Study}

To isolate the contribution of each algorithmic component, we evaluate four cumulative variants across the three RoboMimic tasks: the frozen Naive baseline, LoRA adaptation under request-aligned targets, LoRA adaptation under response-aligned targets, and the complete model incorporating pending-action conditioning. Table~\ref{tab:robomimic_ablation} reports performance across representative delay regimes. Within each random seed, success rates are averaged equally across tasks and delay levels before computing the cross-seed mean and sample standard deviation.

\begin{table*}[!t]
\centering
\caption{RoboMimic ablation results. Rows are cumulative; values are mean success rates $\pm$ standard deviation over three seeds (\%).}
\label{tab:robomimic_ablation}
\small
\setlength{\tabcolsep}{4.5pt}
\begin{tabular}{lcccccc}
\toprule
& \multicolumn{3}{c}{Delay-range averages}
& \multicolumn{3}{c}{Representative delays} \\
\cmidrule(lr){2-4}\cmidrule(lr){5-7}
Method
& Low ($0$--$3$)
& High ($4$--$7$)
& Overall ($0$--$7$)
& $d=0$
& $d=3$
& $d=7$ \\
\midrule
Naive
& $77.4\pm2.4$
& $28.2\pm1.5$
& $52.8\pm1.9$
& $89.1\pm2.7$
& $62.9\pm3.9$
& $13.6\pm0.8$ \\

\quad + LoRA
& $78.2\pm1.3$
& $27.4\pm0.8$
& $52.8\pm1.0$
& $85.6\pm0.4$
& $63.8\pm2.8$
& $12.4\pm2.0$ \\

\quad + response-aligned target
& $61.6\pm2.5$
& $\mathbf{70.2\pm1.2}$
& $65.9\pm1.2$
& $87.1\pm1.0$
& $68.0\pm2.0$
& $55.8\pm1.4$ \\

\quad + pending-action sequence
& $\mathbf{87.8\pm1.4}$
& $\mathbf{70.2\pm1.8}$
& $\mathbf{79.0\pm1.6}$
& $\mathbf{89.3\pm1.3}$
& $\mathbf{87.3\pm1.3}$
& $\mathbf{63.3\pm2.3}$ \\
\bottomrule
\end{tabular}
\end{table*}

Response-aligned targets raise high-delay success from 27.4\% to 70.2\%. Introducing pending-action conditioning increases low-delay success from 61.6\% to 87.8\% while maintaining the high-delay average at 70.2\%. The complete model achieves an overall success rate of 79.0\%, compared with 52.8\% for both Naive and LoRA-only adaptation. These ablations confirm that response alignment drives high-delay robustness, while pending-action conditioning sustains control performance across the evaluated delay range.

\subsection{Real-World Experiments}
\label{sec:real_world_experiments}

\begin{figure*}[!t]
\centering
\includegraphics[width=\textwidth]{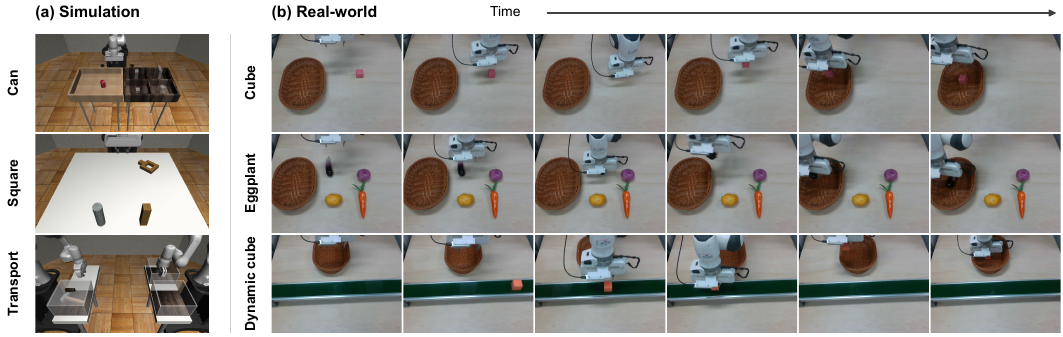}
\caption{Simulation environments and physical execution sequences. (a) RoboMimic simulation tasks: Can, Square, and Transport. (b) Physical pick-and-place tasks: Cube Grasping, Vegetable Grasping with a plastic eggplant, and Dynamic Grasping on a conveyor belt. Each row displays six chronological snapshots from an illustrative trial, with time progressing from left to right.}
\label{fig:real_world_tasks}
\end{figure*}

We evaluate the latency robustness of RAPAC-DP in real-world manipulation across three pick-and-place tasks spanning static and dynamic scenarios. Figure~\ref{fig:real_world_tasks} shows representative execution sequences.

\subsubsection{Experimental Setup and Tasks}

The experimental platform consists of a Franka Panda manipulator with a Franka Hand gripper and two Intel RealSense D435i cameras providing front and wrist views. The local controller communicates wirelessly with a remote inference server equipped with an NVIDIA RTX 5090 GPU. Policy actions are executed at 15~Hz. The policy takes the two RGB views and a four-dimensional state consisting of end-effector Cartesian position and gripper width, predicting relative Cartesian displacements and gripper commands with horizon $H=20$, execution window $K=8$, and 10 DDIM denoising steps.

We evaluate three pick-and-place tasks: \textit{Cube Grasping} with a static cube placed across varied tabletop positions at a fixed orientation; \textit{Vegetable Grasping} using a plastic eggplant; and \textit{Dynamic Grasping} with a cube moving on a linear conveyor belt. Each task requires picking the object and depositing it into a stationary basket without human intervention. Trials terminate upon basket placement or unrecoverable error. Missed grasps, unintended collisions, and dropped objects count as failures, with no mid-trial retries permitted.

For each task, a base Diffusion Policy is trained on 50 demonstration episodes, split into 40 training and 10 validation episodes. RAPAC-DP adapts the frozen base policy by training only the pending-action encoder and LoRA adapters on the same delay-free demonstrations, without requiring additional data collected under delayed execution. RTC and Sync directly employ the corresponding frozen base checkpoints. For all three physical tasks, training delays are sampled uniformly from 0 to 7 control steps.

\subsubsection{Latency Calibration and Asynchronous Execution}

Post-warmup requests characterize baseline communication and inference latencies. We then inject additional latency to obtain nominal execution latencies of 200 and 450~ms. At the 15~Hz control frequency, these time-domain latencies are mapped to discrete delays of $d=3$ and $d=7$, respectively, which determine the scheduled handover step $t+d$ for a request issued at step $t$. During policy inference, the local controller executes committed pending actions and buffers early responses until the scheduled handover, as summarized in Algorithm~\ref{alg:real_latency_handover}.

\begin{algorithm}[t]
\caption{Asynchronous Execution of RAPAC-DP}
\label{alg:real_latency_handover}
\begin{algorithmic}[1]
\Require Policy $\pi$, execution queue $Q$, handover delay $d$
\For{each policy query at control step $t$}
    \State Extract $d$ committed actions $\mathbf{P}_{t,d}$ from queue $Q$.
    \State Dispatch asynchronous request $\mathbf{A} \gets \pi(\mathbf{o}_t, \mathbf{P}_{t,d}, d)$.
    \State Execute committed actions $\mathbf{P}_{t,d}$ toward scheduled handover step $t+d$.
    \State Await response arriving at control step $t_{\mathrm{arr}}$.
    \If{$t_{\mathrm{arr}} \le t+d$}
        \State Buffer plan $\mathbf{A}$ and update $Q \gets \mathbf{A}$ at handover step $t+d$.
    \Else
        \State Hold current robot pose until $t_{\mathrm{arr}}$, then update $Q \gets \mathbf{A}$.
    \EndIf
\EndFor
\end{algorithmic}
\end{algorithm}

RAPAC-DP generates a full $H$-step action chunk aligned with the scheduled handover, whereas RTC executes the frozen base policy with inference-time chunk guidance. Both methods share identical observation and action interfaces, low-level controllers, motion constraints, and latency-injection protocols, with all evaluated delays covered by the trained compensation range.

\begin{figure*}[!t]
\centering
\includegraphics[width=\textwidth]{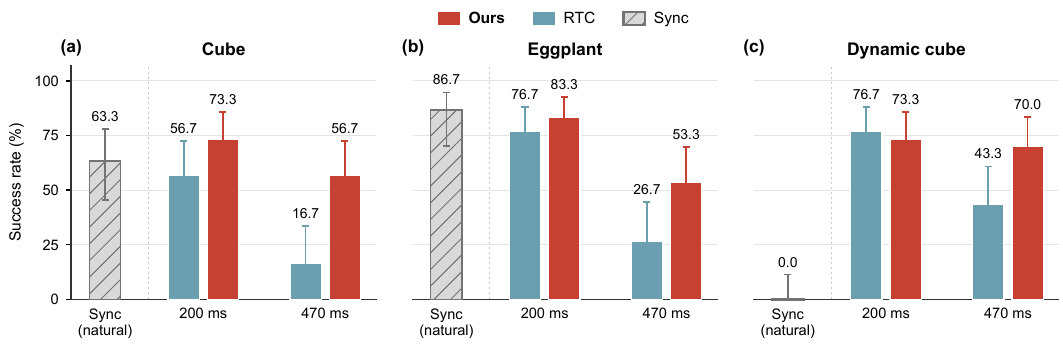}
\caption{Physical pick-and-place success rates across Cube Grasping, Vegetable Grasping, and Dynamic Grasping. RAPAC-DP (Ours) and RTC are evaluated under nominal execution latencies of 200 and 450~ms obtained through latency injection and mapped to discrete delays of $d=3$ and $d=7$, respectively. Sync is evaluated under natural latency as a synchronous reference. Each bar summarizes 30 physical trials for a total of 450 trials; labels indicate success percentages, and error bars denote 95\% confidence intervals.}\label{fig:real_world_success}
\end{figure*}

\subsubsection{Evaluation Protocol and Results}

RTC and RAPAC-DP are evaluated across all three tasks with injected delays resulting in nominal execution latencies of 200 and 450~ms. At a control frequency of 15~Hz, these latencies are mapped to discrete delays of $d=3$ and $d=7$ control steps, respectively. Sync uses the same frozen base policy as RTC and alternates inference with action execution. The robot holds its current pose while waiting for each response, then executes the first $K=8$ actions of the returned chunk at 15~Hz before requesting the next chunk. Each method--task condition comprises 30 physical trials, yielding 360 asynchronous trials and 90 synchronous trials across 450 physical runs in total. A single checkpoint is evaluated per method and task using matched initial configurations and alternated evaluation order, while conveyor speed and starting positions remain constant for Dynamic Grasping. Success rates are reported with 95\% confidence intervals.
  
Figure~\ref{fig:real_world_success} shows the physical success rates under $d=7$. On Cube Grasping, Vegetable Grasping, and Dynamic Grasping, RAPAC-DP achieves success rates of 56.7\%, 53.3\%, and 70.0\%, compared with 16.7\%, 26.7\%, and 43.3\% for RTC. The equal-task average increases from 28.9\% for RTC to 60.0\% for RAPAC-DP, an improvement of 31.1 percentage points.

Under the low-delay condition of $d=3$, RAPAC-DP achieves success rates of 73.3\%, 83.3\%, and 73.3\% across the three respective tasks, compared with 56.7\%, 76.7\%, and 76.7\% for RTC. The equal-task averages are 76.7\% for RAPAC-DP and 70.0\% for RTC, showing that the 6.7 percentage point margin at $d=3$ expands to 31.1 percentage points under the higher delay of $d=7$. Under natural latency, Sync achieves success rates of 63.3\% on Cube Grasping, 86.7\% on Vegetable Grasping, and 0.0\% on Dynamic Grasping. In Dynamic Grasping, the robot pauses while awaiting each inference response while the cube continues moving on the conveyor. The resulting stop-and-go motion prevents the robot from tracking the moving cube.
\section{Limitations}

This work validates RAPAC-DP across both simulation benchmarks and physical manipulation under controlled latency conditions. During training, pending actions are constructed from delay-free demonstrations, while deployment reliably conditions on online policy-generated queues. Future work will extend this framework to stochastic latency jitter and investigate adaptive compensation under time-varying execution delays.

\section{Conclusion and Future Work}

In this paper, we presented RAPAC-DP, a pending-action compensation framework for improving the latency robustness of action-generation policies in imitation learning. RAPAC-DP addresses the mismatch between policy generation and delayed execution by constructing delay-conditioned training samples from delay-free demonstrations. For parameter-efficient adaptation, the base policy remains frozen, while the pending-action encoder and LoRA adapters form the trainable compensation pathway. When desired, this pathway can be explicitly bypassed to recover the frozen base policy exactly.

On RoboMimic, RAPAC-DP achieves a mean success rate of 63.3\% at the maximum evaluated delay, while retaining 81.4\% of its delay-free performance on Kinetix. Across three physical pick-and-place tasks, it increases the mean success rate under severe delay from 28.9\% for RTC to 60.0\%. These results validate response-aligned training with pending-action conditioning as an effective framework for asynchronous robot control under execution delay.

Future work will examine the interaction between inference latency and execution delay, extend RAPAC-DP to large-scale Vision-Language-Action policies and accelerated generative paradigms such as consistency models, and evaluate longer-horizon tasks.

\balance

\bibliographystyle{IEEEtran}
\bibliography{references}

\end{document}